\documentclass[11pt]{article}

\usepackage[T1]{fontenc}
\usepackage[utf8]{inputenc}
\usepackage{lmodern}
\usepackage[margin=1in]{geometry}
\usepackage{microtype}
\usepackage{csquotes}
\usepackage[round,authoryear]{natbib}
\usepackage[hidelinks]{hyperref}
\usepackage{enumitem}

\setlist[enumerate]{leftmargin=*,itemsep=0.35em,topsep=0.4em}

\title{Beyond Epistemia:\\ Epistemic Schizologia and Large Language Models as Techno-Semiotic Machines}
\author{
    \textbf{Federico Cabitza}$^{1,2}$ \\
    \small ORCID: \href{https://orcid.org/0000-0002-4065-3415}{0000-0002-4065-3415} \\
    \small Email: \href{mailto:federico.cabitza@unimib.it}{federico.cabitza@unimib.it}
    \and
    \textbf{Gianluca Colombo}$^{3}$ \\
    \small ORCID: \href{https://orcid.org/0009-0006-6867-0221}{0009-0006-6867-0221} \\
    \small Email: \href{mailto:gianluca.colombo@oneofftech.de}{gianluca.colombo@oneofftech.de}
    \\[2.5ex]
    \small $^{1}$ Department of Informatics, Systems and Communication, University of Milano-Bicocca, Milan, Italy \\
    \small $^{2}$ Digital Health and Wellbeing Center, Fondazione Bruno Kessler, Trento, Italy \\
    \small $^{3}$ Oneofftech-UG (haftungsbeschränkt), Berlin, Germany
}
\date{28 July 2026}

\begin{document}
\maketitle

\begin{abstract}
Quattrociocchi and colleagues warn that the fluent outputs of large
language models may allow linguistic plausibility to substitute for
epistemic evaluation, producing the condition they call \emph{Epistemia}:
the experience of possessing knowledge without undertaking the practices
through which judgment would ordinarily be warranted. This article accepts
that diagnosis but challenges its explanatory framework, which compares an
embodied, socially situated human knower with an isolated generative model
thereby locating epistemic legitimacy in capacities internal to
autonomous agents. Drawing on Carlo Sini's philosophy of practices,
writing, signs, and technics, we propose instead to understand a large
language model (LLM) as a \emph{techno-semiotic machine} that automates a
phase of written semiosis by producing plausible linguistic configurations
from the sedimented archive of human writing.

From this perspective, \emph{Epistemia} is one consequence of a broader
phenomenon that we call \emph{epistemic schizologia}: the socio-technical
cleavage between signs as linguistically accomplished expressions and signs
as moments within socially embedded circuits of interpretation, evidence,
criticism, verification, and responsibility. This cleavage is reinforced by
\emph{eikotic closure}, through which a plausible continuation is presented
with the finality of an epistemic result, and by algorithmic authority and
epistemic self-misrecognition. The relevant unit is therefore not the model alone but the complete
practice in which generated
inscriptions are prompted, interpreted, verified, contested, used, and made
consequential. This reframing preserves the distinction between linguistic
production and responsible understanding while grounding a design programme
centred on inspectable genealogy, contestability, distributed
responsibility, epistemic agency, and the evaluation of hybrid human--AI
practices.
\end{abstract}


\section{A sound warning built on the wrong comparison}

Quattrociocchi and colleagues argue that the increasingly close resemblance between human and machine outputs conceals a more fundamental divergence between the processes through which human beings form judgments and those through which large language models generate text \citep{quattrociocchi2025}. Their analysis proceeds along two closely connected lines: text generation is first described as the completion of stochastic trajectories within a high-dimensional space of linguistic transitions, and the operation of the model is then compared with a human epistemic pipeline divided into seven ostensibly corresponding stages, from which emerge fault lines concerning grounding, parsing, experience, motivation, causal reasoning, metacognition, and value. On this basis, the authors introduce the notion of \emph{Epistemia}, understood as a condition in which linguistic plausibility takes the place of epistemic evaluation and the availability of a persuasive answer creates the impression that the work of judgment has already been performed~\citep{loru2025simulation,quattrociocchi2025}. Epistemia provides a particularly useful point of departure because it names a central transformation introduced by generative AI: the possibility that the recognizable linguistic form of knowledge may circulate independently of the practices through which epistemic warrant is ordinarily established. Our aim is therefore not merely to reject its original formulation, but to situate the phenomenon it identifies within a broader account of technological mediation, semiosis, and epistemic practice.

This diagnosis captures a genuine and consequential danger, since an LLM possesses neither a lived world nor autobiographical experience, intrinsic purposes, moral exposure, or responsibility for the consequences produced by the texts it generates. Its answers may nevertheless exhibit many of the conventional signs by which knowledge is ordinarily recognized---technical vocabulary, syntactic control, orderly exposition, apparent confidence, and conformity to established argumentative genres---even though those answers have not, within the model itself, passed through observation, source criticism, experimental inquiry, testimonial assessment, or accountable verification. When the recognizable form of knowledge is accepted as sufficient evidence that knowledge has been attained, fluency can indeed conceal the absence of the practices through which a claim would normally acquire epistemic warrant.

The principal difficulty, however, concerns not the warning itself but the comparison through which the warning is theoretically articulated. Quattrociocchi and colleagues contrast two entities that have been constituted at markedly different levels of description: on one side stands a fully formed human epistemic agent, already endowed with embodiment, social relations, historical experience, motivations, institutional affiliations, and exposure to practical consequences; on the other stands an isolated computational model, abstracted from the training processes that formed it, the system prompts and retrieval mechanisms that constrain it, the interfaces through which it becomes accessible, the users who interpret its outputs, and the organizations that determine how those outputs will be employed. Because the analytical frame grants the human being the resources of an entire biological, social, and institutional ecology while reducing the artificial system to its generative core, the profound asymmetry subsequently identified between them is not merely discovered by the analysis but, to a considerable extent, established by its premises.

The seven-stage diagram makes this asymmetry particularly evident, because it presents tokenization as the artificial counterpart of perceptual and situational parsing, embeddings as the counterpart of memory and conceptual knowledge, textual context integration as the counterpart of reasoning, and probabilistic prediction as the counterpart of value-sensitive judgment. Quattrociocchi and colleagues are of course explicit that these stages ``diverge sharply in structure, function, and epistemic grounding,'' and the very language of \emph{fault lines} is designed to mark asymmetry rather than equivalence. The difficulty therefore lies at a prior, ontological level rather than at the level of an implicit equivalence claim: constructing a stage-by-stage parallel between a human being and a language model already presupposes that the two processes occupy the same kind of place within comparable epistemic architectures, so that a point-by-point comparison is intelligible at all. Once this presupposition is granted, the ensuing analysis can register asymmetries between corresponding stages, but it cannot ask whether the very idea of a corresponding stage is well formed. From a practice-centred standpoint, tokenization is not an impoverished counterpart of perceptual parsing but a phase of a semiotic operation whose epistemic status is not settled by what happens inside the model, and can be adjudicated only downstream, in the interpretive and institutional uptake of the resulting inscription. The physical movement of a pen does not correspond to the interpretation of the legal argument written with it: both movements may be necessary conditions of a more complex practice, but neither can be treated as its epistemic analogue, and no refinement of the description of the pen would tell us what the argument means, whether it is sound, or by whom it is owed. The difference to which the authors draw attention is therefore real, although the parallel architecture through which they represent it remains conceptually unstable.

The description of generation as a stochastic walk on a graph reinforces the same reduction, even though it serves a legitimate heuristic purpose by emphasizing that LLMs generate probable continuations rather than forming beliefs. As a technical characterization, however, the metaphor is insufficiently discriminating, since the probability assigned to each subsequent token depends on an evolving context and on representations produced through multiple layers of attention, rather than on a fixed transition relation among isolated elements of a vocabulary \citep{vaswani2017}. More importantly, even a computational description that captured the generative mechanism with complete accuracy would not, by itself, determine the epistemological significance of the resulting text, because the causal explanation of an inscription and the practical role performed by that inscription belong to different, although connected, levels of analysis. An account of the physical deposition of ink does not exhaust the nature of writing; in the same way, an account of token generation does not exhaust what generated writing becomes when it enters practices of interpretation, verification, deliberation, and institutional action.

\section{The internalist residue in \emph{Epistemia}}

The deeper limitation of the framework lies in the internalist conception of epistemic legitimacy that organizes its contrasts. Quattrociocchi and colleagues repeatedly locate the decisive difference between human beings and LLMs in capacities that the former are said to possess internally and the latter to lack: perceptual grounding, belief maintenance, causal modelling, metacognitive monitoring, value commitment, and the ability to withhold judgment. These distinctions are indispensable when the question concerns whether an LLM constitutes an epistemic subject comparable to a human being, since no plausible account of present systems warrants such an identification; they are far less conclusive, however, when the question concerns whether, and under which conditions, a generative system may perform a function within a broader practice of knowledge production.

The philosophical difficulty becomes clearer when the human side of this
contrast is treated as though it were grounded in a self-transparent
interiority, already constituted before its involvement with signs,
practices, and technical mediations. Such a premise is difficult to
sustain within a philosophical tradition that, at least since Peirce, has
subjected the alleged immediacy of our access to the internal world to
radical criticism. In \emph{Some Consequences of Four Incapacities},
Peirce rejects the Cartesian image of cognition as originating in an
intuitively accessible interior domain, arguing instead that knowledge of
what occurs within us is inferentially reconstructed from external facts,
that every cognition is conditioned by previous cognitions, and that
thought cannot occur independently of signs
\citep{peirce1868}.\footnote{A contemporary argument developed from a
different, cognitive-scientific perspective can be found in
\citet{chater2018}, who challenges the assumption that thought draws upon
a determinate repository of hidden beliefs, motives, and mental contents,
and proposes instead that the mind continuously improvises interpretations
in response to present circumstances. Chater's thesis does not entail the
Peircean account of cognition as semiosis, but it offers a convergent
critique of the image of a deep, self-contained interior domain from
which judgment would simply emerge.}

Peirce's argument does not diminish the substantial differences between
human beings and computational systems, nor does it imply that the
semiotic activity of a living and embodied subject can be identified with
the operation of a language model. It does, however, undermine any
attempt to locate those differences in a self-sufficient internal domain
that the organism would possess prior to, and independently of, its
relations with the world. Human cognition is itself inferentially
continuous, dependent upon prior cognitions, and mediated by signs;
accordingly, the relevant comparison must be reconstructed at the level
of the different bodies, histories, practices, and circuits of
interpretation through which human beings and technical artifacts perform
distinct functions within epistemic activity\footnote{Here it is worth recalling an evocative passage from Skinner on the differences between humans and machines: ``The real question is not whether machines think but whether men do. The mystery which surrounds a thinking machine already surrounds a thinking man.''~\citep{skinner1969contingencies}}.

Human knowledge has never been produced by an isolated mind confronting an independently given world without technical or symbolic mediation, because even the most elementary epistemic activities depend upon testimony, linguistic classifications, notational systems, instruments, archives, diagrams, laboratories, documentary procedures, and communities capable of criticism and correction. Scientific truth, in particular, is not secured by an individual's possession of a complete internal epistemic loop, but achieved through stabilized practices in which observations are inscribed, methods are rendered inspectable, claims are exposed to refutation, and responsibilities are allocated among persons, instruments, institutions, and procedures. 
Building on Peirce’s critique of introspection, intuition, and signless thought, Sini’s philosophy of practices gives this anti-internalist insight a more explicitly historical and socio-technical articulation. Knowledge is understood not as a disembodied correspondence between an internal representation and an external object, but as a historically constituted activity in which bodies, signs, technical devices, habits, and institutions jointly shape what may count as an object, an observation, a reason, or a warranted assertion~\citep{sini1996,sini2000}.

Such a position does not imply that every technically mediated assertion qualifies as knowledge, nor does it erase the decisive differences between human activity and machine operation. It implies, more narrowly and more rigorously, that the absence of belief, consciousness, or intrinsic purpose in an artifact cannot, by itself, determine the epistemic status of the sociotechnical practice in which that artifact is embedded. A telescope does not perceive as a human observer perceives, a database does not remember through autobiographical recollection, and a statistical procedure does not understand the variables over which it operates; nevertheless, each may perform an indispensable epistemic function when its operation is incorporated into practices that specify admissible inputs, establish procedures of control, expose results to contestation, and assign responsibility for interpretation and use. The pertinent question is therefore not whether the artifact internally reconstructs the complete architecture of human judgment\footnote{``Epistemia is, instead, an \emph{architectural phenomenon} that arises whenever generative systems are inserted into
epistemic workflows while lacking \emph{internal machinery} for
reference, verification, or belief maintenance'' (our emphasis). ``Epistemia [\ldots] arises from \emph{architectural features of generative
systems}, and therefore persists even when outputs are accurate, calibrated, or behaviorally aligned.'' (our emphasis)~\citep{quattrociocchi2025}.}, but which operation has been delegated to it, according to which constraints that operation is performed, and at which points interpretation, verification, decision, and accountability remain situated.

The human side of the proposed fault lines is, moreover, described in terms that risk idealizing capacities which human beings possess only unevenly and exercise with considerable fallibility. People frequently reason from correlations without identifying causal mechanisms, misread social situations, express poorly calibrated confidence, rationalize commitments formed for non-epistemic reasons, and fail to suspend judgment when the available evidence would warrant restraint. Epistemic reliability often arises not because individual cognition spontaneously overcomes these limitations, but because practices compensate for them through written records, replication, adversarial review, procedural checks, division of cognitive labour, and institutional mechanisms designed to expose error. To characterize human beings by reference to the corrective resources of the entire environment in which they act, while characterizing LLMs exclusively by reference to the limitations of their internal architecture, is therefore to compare the situated achievements of a sociotechnical practice with the isolated mechanism of one of its possible components. The resulting contrast identifies genuine differences, but it does not yet provide an adequate theory of how epistemic authority is produced, distributed, and sometimes misplaced within hybrid human--machine arrangements.

\section{The language model as a techno-semiotic machine}

A Sinian account begins from a different philosophical framework, according to which the human being is not first constituted as an autonomous cognitive subject and only subsequently furnished with signs, instruments, and external supports; rather, distinctively human forms of thought emerge historically through the technical and semiotic exteriorizations by which bodily practices are preserved, transformed, and transmitted. Alphabetic writing, tables, diagrams, archives, and computational media do not merely convey a cognition that has already taken shape elsewhere, within an independently constituted mind, but reorganize the conditions under which something may be perceived, remembered, demonstrated, compared, or constituted as an object of knowledge \citep{sini1994,sini1997,sini2009}. If this thesis is accepted---and it is supported by substantial philosophical, philological, and anthropological traditions---generative artificial intelligence should be situated primarily within the history of writing and inscription practices, and only subsequently compared with the capacities of human cognition, whose allegedly self-sufficient character is itself the product of a particular technological and philosophical history.

The notion of a \emph{techno-semiotic machine} is useful in this context because it characterizes the large language model without assimilating it either to a human mind or to an inert computational instrument. Such a machine is situated downstream from an immense archive of human inscriptions, whose forms have been statistically incorporated into its parameters, and upstream from the acts of interpretation, verification, use, and institutionalization through which its outputs may acquire significance and practical force. It neither retrieves intact propositions from an internal library nor produces meaning \emph{ex nihilo}; instead, it incorporates recurrent structures of writing---including genres, argumentative patterns, registers, classifications, styles, conventional associations, and modes of textual organization---and reactivates them as linguistic configurations adapted to a locally specified context. In this precise and deliberately restricted sense, the model automates one phase of written semiosis: not the entire production of meaning, which also requires interpretation and use, but the generation of fluent, context-sensitive, and conventionally appropriate textual forms.

This restricted characterization can be clarified by considering an
apparently contrary case. Certain twentieth-century avant-garde
experiments, most notably within Dada, sought deliberately to disrupt
established semantic conventions through arbitrary juxtaposition,
fragmented syntax, chance procedures, and compositional strategies
intended to frustrate habitual expectations of coherence. Yet these
experiments did not suspend semiosis, because their provocations became
intelligible precisely insofar as readers continued to mobilize inherited
interpretive habits, literary genres, cultural expectations, and symbolic
associations in attempting to make sense of them.

The deliberate disturbance of established codes therefore did not abolish
meaning; rather, it enlarged the openness of the work and transferred a
greater share of semiotic labour to the interpreter
\citep{eco1989,forcer2015}. Even an inscription arranged to resist
interpretation continues to solicit interpretants, whether it is received
as parody, protest, absurdity, formal experiment, or critique of the very
conventions whose operation it disrupts. The aspiration to produce pure
unintelligibility thus reveals the persistence, or even the inertia, of
sign relations: once an inscription enters a historically constituted
semiotic environment, it cannot prevent itself from becoming available
for further interpretation. Peirce's fourth incapacity \citep{peirce1868} acquires an
unexpected resonance here: we have ``no conception of the absolutely
incognizable.''  Complete semantic rupture remains unattainable, not
because every text possesses a determinate meaning concealed within it,
but because every trace is taken up within an already operative network
of signs and interpretive practices.

The large language model occupies, in this respect, the opposite pole of
the avant-garde experiment. Rather than attempting to interrupt inherited
regularities, it statistically incorporates, amplifies, and recombines
them, producing linguistic configurations whose intelligibility depends
upon their conformity to sedimented genres, associations, and patterns of
discourse. Its outputs are intelligible not because the model possesses
an internal understanding of what it says, but because it reactivates the
historically accumulated regularities through which human semiosis has
stabilized recognizable forms of meaning. The LLM and the Dadaist
experiment thus belong, although in opposing ways, to the same technical
history of writing: one exposes the difficulty of escaping inherited sign
relations, while the other derives its efficacy from reproducing them at
scale. Both cases show that meaning is generated neither by an isolated
mind nor by an isolated text, but through the continuing operation of
historically constituted practices of inscription and interpretation.

Recent work in the philosophy of technology has similarly argued that generative AI requires a system-level reconceptualization because it produces semantic artefacts that alter socio-technical relations \citep{bisconti2024}. We share the insistence that the social significance of generated language cannot be inferred from the model in isolation. We differ, however, from accounts that describe generative systems themselves as poietic epistemic agents: on our view, their capacity to generate consequential inscriptions is a form of semiotic efficacy, but not by itself evidence of understanding, commitment, or responsibility. These remain properties and achievements of the wider practice, distributed asymmetrically among its human and institutional participants.

This characterization is intended to avoid two opposed, although frequently complementary, errors that continue to organize much of the debate concerning generative AI. Anthropomorphism must be rejected because the model neither inhabits a lived world nor assumes responsibility for the signs it produces, and because it lacks the biographical, practical, and normative horizon within which human assertions become commitments for which someone can be called to account \citep{benderkoller2020}. Instrumental reductionism must also be rejected, however, because the inscriptions produced by the model are not merely inert strings whose significance begins only when an external observer arbitrarily assigns meaning to them; they carry sedimented semantic orientations, evoke conventional expectations, invite particular inferences, enter documents and decisions, and may acquire the force of authoritative discourse when institutional conditions allow them to do so. Meaning is therefore not enclosed within the model as an internal possession, but neither is it simply appended after generation: it emerges through the semiotic circuit in which a linguistic configuration is requested, produced, interpreted, tested, revised, accepted or rejected, and eventually made consequential. In terms derived from Peirce and developed by Sini, the life of the sign resides in the series of interpretants through which it is repeatedly taken up, as well as in the practices and material arrangements that sustain this process \citep{peirce1931,sini1978}.

A recent Peircean account likewise distinguishes LLMs' sophisticated representational capacities from intentionality, grounding the latter in embodied forms of semiosis that current systems do not reproduce \citep{garciavaldecasas2026}. Our account is compatible with this restriction but addresses a different question: not whether the model itself is intentional, but how non-intentional generative operations can transform the conditions under which human assertions become meaningful, authoritative, and actionable.

The appropriate unit of analysis is consequently not the model considered in isolation, but the entire circuit within which generated language becomes operative:

\begin{quote}
\emph{prompt $\rightarrow$ model and technical assemblage $\rightarrow$ generated inscription $\rightarrow$ interpretation $\rightarrow$ verification $\rightarrow$ use $\rightarrow$ consequence and responsibility.}
\end{quote}

The sequence should not be understood as a rigidly linear pipeline, since interpretation may modify the prompt, verification may require renewed generation, and practical consequences may retrospectively alter the criteria by which an output is assessed; it nevertheless makes visible the plurality of operations and responsibilities that disappear when the model alone is treated as the bearer of epistemic agency. The prompt establishes a discursive situation and selects the task to be performed, while the technical assemblage constrains the sources, tools, memories, and policies available to the model; the generated inscription then becomes the object of interpretation and possible verification, after which it may be incorporated into a decision, a document, or an institutional procedure whose consequences must be attributed to identifiable human and organizational actors. The model performs an indispensable operation within this circuit, but the epistemic status of the result cannot be inferred from that operation alone.

This change in analytical focus is therefore not merely terminological, because generative systems deployed in settings of practical consequence are rarely reducible to a bare language model. They usually include retrieval mechanisms, external databases, domain-specific tools, system prompts, memory components, policy and social constraints, user interfaces, procedures of human review and reinforcement, and organizational rules and conventions governing whether and how generated content may be acted upon. 
Quattrociocchi and colleagues do not overlook these components; they explicitly consider retrieval-augmented generation, tool use, external memory, and reinforcement learning with human feedback, and conclude that such mechanisms should be treated as ``partial mitigations rather than epistemic solutions'' whose addition can even ``intensify Epistemia'' by increasing the persuasive authority of outputs while leaving responsibility diffuse. 

This verdict, however, is itself a symptom of the internalist premise identified above rather than an independent finding: because augmentation cannot install beliefs, purposes, or metacognition within the model, it is judged epistemically inert, and its evident capacity to reorganize the surrounding practice is thereby made invisible. A practice-centred framework reaches a different assessment, not by denying that these components fail to confer consciousness or intrinsic purposes upon the model---they plainly do not, and none transforms the machine into a human-like subject---but by observing that it is precisely at the level of practice, and not at the level of an internal state, that such components modify the epistemic quality of what is done, reconnecting linguistic generation with documentary evidence, procedural checks, explicit uncertainty, and accountable decision-making.

The difference can be illustrated by comparing a clinician who consults a general-purpose language model in isolation with one who uses the same generative model through a system that retrieves the relevant passages of an accredited clinical guideline, distinguishes those passages visually from the model's synthesis, displays conflicts among sources, and requires the clinician to verify and countersign any assertion before it enters the medical record. The generative mechanism may be substantially identical in the two cases, yet the epistemic configuration is not, because the second arrangement preserves the documental genealogy of the claim, creates opportunities for contestation, and locates responsibility at identifiable points in the clinical workflow. A framework that makes epistemic legitimacy depend primarily on whether the model internally possesses capacities analogous to human understanding cannot adequately describe this difference, whereas a practice-centred framework can explain both why the model remains non-human and why the systems built around it may nevertheless differ profoundly in reliability, accountability, and epistemic value.

\section{From \emph{Epistemia} to epistemic \emph{schizologia}}

The concept of \emph{Epistemia}, which Quattrociocchi, Capraro, and
Perc define as the condition in which linguistic plausibility substitutes
for epistemic evaluation, identifies a genuine and consequential danger
\citep{quattrociocchi2025}. Generative systems can provide users with
the experience of possessing an answer without requiring them to traverse
the practices through which a warranted judgment would ordinarily be
formed, so that a fluent and apparently complete formulation may come to
stand in for inquiry, evidential assessment, criticism, and responsible
deliberation. We accept this diagnosis, but regard it as describing only
one manifestation of a more extensive phenomenon, which cannot be
adequately understood if attention remains confined to the epistemic
capacities allegedly present or absent within the model.

We call this broader phenomenon \emph{epistemic schizologia}, from the
Greek \emph{schizein}, to divide, and \emph{logos}, understood as
discourse, account, or reasoned formulation.\footnote{Our use of
\emph{schizologia} is conceptually distinct from Gilles Deleuze's
\emph{Schizologie}, the title of his 1970 preface to Louis Wolfson's
\emph{Le Schizo et les langues}. Deleuze examines Wolfson's singular
procedures of linguistic transformation, whereas \emph{epistemic
schizologia}, as defined here, concerns the socio-technical separation
of generated epistemic discourse from the practices of interpretation
and verification through which it could acquire warranted authority.
The shared etymological construction implies no claim of conceptual
derivation or continuity \citep{deleuze1970}.} In its simplest semiotic
formulation, \emph{epistemic schizologia} names the cleavage between the sign
considered as a linguistic expression and the sign considered as an
episode within a socially embedded circuit of interpretation. On one
side stands the generated utterance, which may exhibit the vocabulary,
argumentative structure, qualifications, and institutional tone of
reasoned knowledge; on the other stand the practices through which that
utterance would be related to evidence, interpreted by multiple
participants, exposed to criticism, revised, and made answerable for its
consequences.

This distinction shifts the analysis away from the contrast between a
human subject who allegedly knows and a machine that does not. Large
language models do not merely produce conclusions while omitting an
internal reasoning process that human beings would otherwise perform;
they reproduce the visible morphology of accomplished reasoning
itself---definitions, distinctions, explanatory transitions,
counterarguments, qualifications, and apparently warranted conclusions.
The resulting discourse may therefore resemble the product of epistemic
labour even when it is no longer accompanied by a recoverable relation
to the observations, documents, disagreements, interpretive decisions,
and responsible acts through which such labour would ordinarily be
conducted.

This account does not imply that generated discourse is semantically
empty. From a Derridean perspective, Gunkel characterizes LLMs as
\emph{différance engines}: systems in which linguistic significance
emerges through differential relations, temporal deferral, trace, and
iterability rather than through the prior intention of a sovereign
speaking subject \citep{gunkel2026differance}. We accept that generated
inscriptions may become meaningful through these relations and through
their interpretive uptake. \emph{Epistemic schizologia} concerns a different
separation: not the absence of meaning from generated language, but the
detachment of meaningful and epistemically recognizable discourse from
the evidential, interpretive, and institutional practices through which
its particular assertions could acquire warrant.

By extending Peirce's vocabulary, this semiotic cleavage may be described
as a \emph{de-indexicalization of warrant}. The symbolic conventions and
iconic features through which a text is recognized as knowledgeable
discourse remain fully operative: the terminology is appropriate, the
argument appears coherent, and the genre conforms to established
expectations. The indexical relations through which its particular
assertions could be traced to documents, observations, sources,
experiments, and acts of inquiry, however, may become attenuated,
concealed, or entirely unavailable \citep{peirce1931}. What circulates
is consequently not an insignificant sequence of words, but a
semiotically powerful utterance whose marks of epistemic authority have
become separable from the practical genealogy that could warrant them.

A closely related Peircean analysis characterizes basic language models as operating within a ``hall of mirrors,'' reproducing linguistic surfaces without adequate indexical grounding or participation in socially mediated epistemology, and asks whether memory, tools, and mediated interaction with the world might enable newer systems to become genuine interpretants \citep{manheim2026}. This diagnosis converges with our account in identifying the separation of symbolic competence from indexical and social relations. The difference concerns both the unit of analysis and the normative direction of the argument. The question posed by \emph{epistemic schizologia} is not primarily whether technical augmentation can turn the artificial system itself into a Peircean interpretant. It is whether the generated inscription remains embedded in a socio-technical circuit in which human and institutional interpretants can recover its objects, inspect its transformations, contest its claims, and bear responsibility for its consequences. Retrieval, memory, and tool use matter on this account not because they necessarily confer epistemic subjecthood upon the model, but because they can reconfigure the warranting semiosis in which its outputs are taken up.

Hallucination is one conspicuous manifestation of this separation,
because an invented fact or source makes the absence of an adequate
indexical relation immediately visible; factual falsity is not, however,
the defining feature of \emph{epistemic schizologia}. A generated statement may
be factually correct and nevertheless remain schizological when it
presents a contested interpretation as settled, suppresses the
conditions under which the relevant evidence was produced, conceals the
transformations applied to source materials, or adopts the rhetoric of
warranted assertion without rendering its warranting practices
inspectable. The problem resides not only in whether the linguistic
configuration corresponds to what is the case, but in whether it remains
connected to the interpretive and institutional circuit within which
that correspondence could be examined, contested, and responsibly
affirmed.

\emph{Epistemic schizologia} is sustained through two complementary movements,
which operate from opposite sides of the human--machine relation. The
first proceeds from system design toward the user and consists in the
engineered presentation of generative discourse as a complete and
self-sufficient answer. We call this movement \emph{eikotic closure},
from the Greek \emph{eikos}, which designates what is likely, plausible,
or reasonable. Plausibility is indispensable to many forms of inquiry,
which necessarily proceed through defeasible hypotheses and provisional
conclusions; eikotic closure occurs, however, when a plausible
linguistic continuation is given the interactional finality of an
epistemic result, while uncertainty, evidential plurality, unresolved
disagreement, and the possibility of further interpretation recede from
view. The polished response closes, at the level of the interface, a
process that remains open at the level of warrant.

This closure is not simply an accidental property of probabilistic
generation, because it is reinforced by deliberate choices concerning
interface form, interactional rhythm, default response structure,
provenance, citation, uncertainty, and opportunities for contestation.
A system may present its output as a provisional synthesis whose sources
and inferential transformations remain visible, or it may present the
same output as an authoritative answer that appears to require no
further inquiry. The underlying model may be unchanged, yet the semiotic
and epistemic character of the practice differs profoundly. What is
engineered is not only the production of linguistic configurations, but
the manner in which those configurations invite recognition and
acceptance.

The second movement proceeds from users and institutions toward generated
discourse, whose schizological condition may be overlooked or actively
accepted. Algorithmic authority~\citep{milella2026perceiving} arises when credibility is attributed to
an utterance because of the technical system that produced it, the
institutional environment in which it appears, or the stylistic
competence with which it is expressed, rather than because its
warranting genealogy has been adequately inspected. Epistemic
self-misrecognition accompanies this attribution when users mistake
access to a persuasive formulation for possession of the grounds that
would justify its acceptance.

This reception-side dynamic is related to the Wittgensteinian account of the ``bewitchment'' that arises when conformity to familiar linguistic practices is taken as evidence of communicative understanding \citep{bottazzi2025}. \emph{Epistemic schizologia} extends that insight from the attribution of understanding to the attribution of warrant: the user encounters not only an apparently understanding interlocutor, but an apparently completed discourse whose linguistic form invites recognition as the outcome of inquiry.

Such misrecognition, called also ``illusion of knowledge''~\citep{loru2025simulation} and ``feeling of knowing''~\citep{quattrociocchi2025}, should not be reduced to individual credulity or
deliberate self-deception, because it is encouraged by broader habits of
cognitive and cultural consumption. The expectation that information
should be immediate, fluent, personalized, and frictionless disposes
users to treat the completed answer as the natural unit of knowledge,
while the slower practices of source comparison, interpretation,
criticism, and revision come to appear as avoidable costs. Under
conditions of epistemic fatigue, institutional time pressure, or
consumerist habituation to instant satisfaction, warrant-de-indexicalized
utterances may consequently appear sufficient not because their
limitations are invisible in principle, but because the practices
required to overcome those limitations have been progressively
devalued.

\emph{Epistemia} emerges within this broader schizological condition,
but it should not be identified with it. \emph{Epistemic schizologia} names the
cleavage between linguistic configuration and socially embedded
semiosis; eikotic closure names the designed movement through which this
cleavage is obscured beneath the apparent completeness of a plausible
answer; algorithmic authority and epistemic self-misrecognition describe
the complementary movement through which users and institutions accept
that answer as epistemically sufficient. \emph{Epistemia}, in
Quattrociocchi and colleagues' sense, is one characteristic consequence
of these converging movements: the condition in which plausible
discourse is consumed as a substitute for the practices of judgment.

This reconstruction preserves the force of their warning while rejecting
the internalist assumptions that limit its explanatory reach. The deeper
problem is not that the model fails to reproduce an epistemic process
located inside a human subject, because human cognition itself has never
been confined to such an interior space. Knowledge is extended across
bodies, signs, instruments, documents, institutions, and communities of
interpretation, whose activity continuously connects assertions with
evidence, consequences, and further interpretants. From this
practice-centred perspective, the relevant opposition is not between an
internally grounded human cognition and an internally ungrounded
artificial one, but between semiotic circuits that preserve the relations
among expression, interpretation, evidence, and responsibility, and
circuits in which those relations have been severed or rendered
invisible.

Addressing \emph{epistemic schizologia} therefore requires more than teaching
users to distrust fluent outputs or attempting to furnish models with
simulated analogues of belief, metacognition, and epistemic intention. It
requires that the two sides of the cleavage be brought back into relation:
generated linguistic configurations must be reinserted into practices of
semiosis in which their sources can be recovered, their transformations
examined, their claims contested, and their consequences assigned to
responsible actors, that is \emph{warranting semiosis}\footnote{This phrase translates what Quattrociocchi and colleagues call ``labour of judgment'' from their internalist vocabulary---something accomplished inside an epistemic agent---into our Sinian and Peircean vocabulary, in which judgment is constituted through extended practices involving signs, objects, interpretants, documents, instruments, institutions, and responsible actors.}. In Peircean terms, the task is not to arrest the sign
at the moment of generation, treating the utterance as a completed unit
of knowledge, but to restore it to the potentially unbounded movement of
interpretation through which signs are assembled, taken up by multiple
interpretants, corrected, and institutionally tested.

\section{Designing practices rather than simulating agents}

The practical consequences of the two frameworks diverge most clearly at the level of design. A theory centred upon deficits internal to the model tends to ask how those deficits might be compensated, whereas a practice-centred account asks how the complete human--machine configuration should be organized so that interpretation, verification, and responsibility remain possible. Quattrociocchi and colleagues themselves advance a substantial socio-technical programme under three headings---epistemic evaluation beyond surface alignment, epistemic governance beyond behavioural alignment, and epistemic literacy beyond critical thinking---and many of the measures they advocate, including provenance, the communication of uncertainty, human review, abstention, disclosure of what a system did \emph{not} do, and calibrated participation by users, are compatible with, and in some cases directly required by, the reframing proposed here.

The disagreement bears not on these recommendations but on the explanatory grammar within which they are inscribed. In their framework, socio-technical measures function as compensations for capacities that the model lacks internally, so that the surrounding practice is called upon to supply what the system cannot provide by itself. In a Sinian reframing, by contrast, the practice is not compensatory but constitutive: it is the site at which epistemic status is produced in the first place, whether the operations delegated to a component are performed by a human being, an instrument, or a language model. This reframing therefore retains the measures Quattrociocchi and colleagues propose but places them within a broader design programme directed toward the practices through which generated signs become credible, actionable, and institutionally consequential.

The first commitment concerns \emph{the inspectability of genealogy}, since a generated claim should not be presented as though it emerged without historical or documentary antecedents. At the level of the interface, generated synthesis should remain distinguishable from retrieved evidence, while users should be able to recover the relevant sources, passages, transformations, and unresolved disagreements that contributed to the answer. This requirement exceeds the addition of decorative citations or generic source lists, which may create an appearance of transparency without allowing the relation between a claim and its evidential basis to be inspected. Provenance, properly understood, restores the documentary chain through which an assertion can become contestable and through which those who rely upon it can determine what was retrieved, what was inferred, what was reformulated, and what remains unsupported.

The second commitment concerns \emph{the preservation of contestation and of the negative}, because epistemic value does not arise solely from the production of complete and helpful answers but also from the capacity to disclose what resists completion. An interface should therefore permit, and visibly represent, abstention, missing information, counterexamples, conflicting interpretations, evidential gaps, and cases in which the available materials do not justify a determinate conclusion. Deliberate friction may be required when the consequences of an error are serious or when an apparently coherent synthesis conceals significant disagreement \citep{natali2025friction}. Difficulty is not valuable in itself; rather, the interruption of fluent completion can restore the user's attention to the conditions under which acceptance would be warranted. The pause, the unresolved alternative, and the resistant datum are not defects that every interface should eliminate: in appropriately designed epistemic practices, they function as resources for judgment.

The third commitment concerns \emph{the representation of distributed authorship and responsibility}, since no consequential generative system has a single author in the traditional sense. The resulting text is produced through a sequence of contributions that may include the construction of training corpora, model development, system prompting, retrieval design, user prompting, selection among alternatives, revision, verification, authorization, and practical implementation. Although these contributions are not equivalent and should not be collapsed into an undifferentiated notion of co-authorship, they jointly shape the content and effects of the generated inscription. Recent virtue-epistemological work on AI-assisted authorship similarly relocates human epistemic competence from direct sentence production to higher-order acts of architectural design, dialogical refinement, verification, and synthesis \citep{rodrigues2026}. That model clarifies how curatorial activity may preserve authorship and epistemic credit; our concern is broader, because responsibility must also remain reconstructible across the organizational and technical chain through which a generated text is authorized and made consequential. Interfaces and procedures should therefore neither attribute agency imaginatively to the model nor dissolve accountability into the vague assurance that a ``human'' remains somewhere in the loop.

The fourth commitment concerns \emph{the preservation of the user's epistemic agency}, which requires more than retaining a formal right to approve or reject a machine-generated proposal. In settings where generated content may influence consequential decisions, users should be supported in comparing claims with evidence, revising both machine and human formulations, requesting alternative interpretations, and articulating reasons for acceptance, rejection, or suspension of judgment. The objective is not to impose permanent distrust, which would make the system unusable and disregard cases in which reliance is appropriate \citep{cabitza2026calibrating}, but to cultivate calibrated participation by ensuring that users can recognize when verification is required and retain the practical competence, access, and authority necessary to perform it.

This requirement also implies that efficiency cannot be treated as the sole measure of successful interaction. A system that accelerates decision-making while progressively eroding the user's capacity to inspect evidence may improve immediate throughput at the cost of longer-term epistemic dependence. Interfaces should consequently be assessed not only by whether they reduce effort but also by how they redistribute it: whether they remove unnecessary clerical labour, preserve attention for difficult judgments, or instead encourage users to accept completed formulations whose grounds they are no longer equipped to reconstruct. Epistemic agency is therefore not synonymous with manual control over every operation, but with the maintained ability to understand the relevant limits of delegation and to intervene effectively when those limits are reached.

The fifth commitment concerns \emph{the evaluation of the hybrid practice}, since neither model accuracy nor similarity between human and machine answers is sufficient to determine whether a generative system improves the production of knowledge. Evaluation should include the ability of users to recover sources, detect errors, identify unsupported inferences, abstain appropriately, revise initial judgments, and calibrate confidence. It should also examine whether the arrangement supports learning \citep{vicente2025mentoring}, distributes workload sustainably, preserves relevant expertise, and makes responsibility intelligible when an error occurs. The principal outcome is therefore not the isolated performance of either the human or the model, but the quality and resilience of the joint epistemic practice, especially under conditions of uncertainty, novelty, disagreement, and institutional pressure.

Prompting and alignment, considered within this framework, also become epistemically significant design objects rather than merely technical techniques for obtaining desirable responses. A prompt is not a neutral container for an independently formed request: it assigns roles, establishes purposes, defines what counts as relevant, and stages the discursive relation within which the generated answer will appear. Alignment is similarly more than behavioural correction applied to an otherwise neutral model. It inscribes selected normative preferences, prohibitions, and expectations into the probabilities of discourse, and the selection and aggregation of human feedback determine which perspectives become dominant, which alternatives recede, and which forms of uncertainty or refusal can be expressed \citep{gonzalezbarman2025,ouyang2022}.

System prompts, default instructions, tonal conventions, refusal policies, retrieval priorities, and response formats therefore help determine not only what the system says but also how its assertions present themselves to the user---as advice, evidence, suggestion, explanation, command, or apparently neutral description. Their design and audit cannot be separated from the epistemology of the interaction, because a stylistic decision that makes an answer sound measured and authoritative may alter reliance even when its propositional content remains unchanged, while a response format that juxtaposes evidence, inference, and uncertainty may preserve distinctions that fluent prose would otherwise conceal \citep{hildebrandt2015,ouyang2022}. The normative question is consequently not exhausted by whether the model's outputs conform to an externally specified set of acceptable behaviours, but extends to the forms of attention, interpretation, and responsibility that the entire interaction encourages.

\section{Conclusion}

Quattrociocchi, Capraro, and Perc are right to reject the inference from linguistic fluency to human-like judgment, and they are equally right to observe that generative interfaces can make plausibility feel like knowledge, thereby obscuring or displacing the labour of verification and judgement~\citep{quattrociocchi2025}. Their account identifies a serious epistemic risk, particularly in environments where generated answers are consumed without access to their documentary antecedents or without sufficient opportunities for critical examination. The limitation of their framework arises not from this warning, but from the decision to explain the phenomenon by opposing two autonomous epistemic pipelines and locating legitimacy primarily in capacities internal to the agent.

As we have argued, this comparison places unlike units at the same level: the human side includes the resources of embodiment, history, sociality, institutions, and corrective practices, whereas the artificial side is represented largely through the internal mechanism of the generative model. The result is an analysis that idealizes human cognition and treats the model as though its significance could be derived from its architecture alone. This is not a matter of overlooking the socio-technical dimension: Quattrociocchi and colleagues devote a substantial part of their paper to evaluation, governance, and literacy in hybrid human--AI systems, and many of the measures they propose are entirely congenial to a practice-centred view. The point is rather that within their explanatory framework the practice is called upon to \emph{compensate for} what the model internally lacks, whereas on the account defended here the practice is the site at which epistemic status is \emph{constituted} in the first place. The seven fault lines reveal genuine asymmetries, but those asymmetries do not, by themselves, determine the epistemic status of the hybrid systems in which language models are used.

The Sinian approach provides a more robust framework because it begins at the level where generative artificial intelligence first becomes meaningful: the historically constituted practice of signs. Since human knowledge has always depended upon technical and semiotic exteriorizations, the presence of an artifact that lacks consciousness or belief does not place its operations outside epistemic life; rather, it requires us to determine what part of the practice has been delegated, how the resulting inscriptions are interpreted and verified, and where authority and responsibility are located. This perspective can acknowledge every relevant difference between a living human being and a language model without assuming that the simulation of an autonomous epistemic agent is the only philosophically significant issue.

It can therefore explain why a machine that neither understands in the human sense nor bears responsibility for its outputs may nevertheless transform the conditions of understanding, authorship, authority, and truth. By automating the generation of culturally recognizable linguistic forms, the model changes the distribution of semiotic labour: it alters who or what produces the first formulation of a claim, how quickly alternative texts can be generated, which discursive conventions become dominant, and how readily an assertion may acquire institutional visibility. These transformations remain philosophically and politically consequential even when no artificial subject emerges behind them.

The Sinian framework also remains applicable as generative systems acquire new modalities, retrieval mechanisms, external tools, persistent memories, and institutional functions, because its object is not a fixed computational architecture but an evolving configuration of bodies, signs, technical operations, documentary resources, and social practices. A theory that defines the relevant difference exclusively through the present limitations of language models risks becoming obsolete as those limitations are redistributed across larger technical assemblages; a practice-centred theory can instead examine how each new capacity modifies the complete circuit of production, interpretation, validation, and consequence.

The seven fault lines should therefore be reformulated not as seven demonstrations that machine-generated language necessarily stands outside epistemic life, but as seven asymmetries, manifestations of what we called \emph{epistemic schizologia}, that must be recognized and governed within hybrid practices. The absence of sensory grounding, intrinsic motivation, autobiographical memory, causal understanding, metacognitive awareness, and value commitment remains decisive when assigning roles and responsibilities, but it does not relieve designers and institutions of the obligation to examine how these absences are compensated, concealed, or amplified by the surrounding configuration. The relevant question is not whether the large language model reproduces the human path to judgment, since it plainly does not, but what forms of human judgment are cultivated, displaced, weakened, or made newly possible when part of linguistic production is delegated to an automated exteriorization of written culture.

From the perspective of design, this question is more demanding than determining whether an output resembles the product of human cognition, because it requires the evaluation of interfaces, workflows, institutions, incentives, competencies, and relations of accountability alongside the model itself. It is also more useful, since the societies in which generative systems are being introduced do not need to decide merely whether machines know; they must determine how practices of inquiry, judgment, verification, and epistemic warrant should be reorganized when machines can produce, with unprecedented speed and fluency, the signs through which knowledge is ordinarily recognized.

We have defined \emph{epistemic schizologia} as the technically and socially sustained cleavage between epistemic logos\footnote{Logos here identifies with considerable precision the pole that has become technically detachable with the advent of LLMs. Its semantic range includes speech, statement, discourse, account, argument, and reasoned formulation. The LLM produces precisely such a logos: not a bare sign, but an articulated account that may include explanations, distinctions, objections, evidence-like elements, and conclusions. What \emph{schizologia} names is the separation of this epistemically recognizable \emph{logos} from the semiosis in which it would ordinarily acquire warranted force---that is, from the historically sedimented practices through which, for roughly twenty-five centuries in the Western tradition, \emph{logos} has denoted both reasoned human discourse and the discursive activity of reasoning itself.}---the plausible linguistic configuration that displays the recognizable form of knowledge---and the open circuit of semiosis through which that configuration would be related to its objects, interpreted by multiple interpretants, tested against evidence, contested, revised, and made answerable for its consequences. But, beyond the mere task of conceptual definition, our principal concern lies with the deontological implication that follows directly from the preceding argument. \emph{Epistemic schizologia} is neither a purely computational defect nor a
purely psychological illusion, since one side of the cleavage is
technically designed and engineered, while the other is socially
recognized, institutionally reinforced, and psychologically sustained. Epistemia, as characterized by Quattrociocchi and colleagues, offers a valid diagnosis of one salient consequence of this condition, but it lacks the phenomenological and semiotic depth required to identify the broader cleavage from which that consequence arises.

Responsibility must extend across the entire configuration:
to those who design how generated signs are presented, to the
organizations that determine how they may be used, and to the practices
through which users are enabled---or discouraged---to interpret and
verify them. The relevant design question is consequently not how to make
the model resemble an autonomous epistemic subject, but how to construct
socio-technical arrangements capable of reconciling what \emph{schizologia} separates: plausible linguistic expression and the living practices of
interpretation, inquiry, criticism, and responsibility.

\section*{Statements and Declarations}

\noindent\textbf{Funding.} The authors declare that no funds, grants, or other support were received during the preparation of this manuscript.

\noindent\textbf{Competing interests.} The authors have no relevant financial or non-financial interests to disclose.

\noindent\textbf{Author contributions.} FC conceived the original idea and prepared the preliminary draft of the manuscript. FC and GC subsequently developed the argument, revised and expanded the text, and contributed equally to the preparation of the final version. Both authors reviewed and approved the submitted manuscript and accept responsibility for its content.

\noindent\textbf{Ethics approval.} Not applicable. This article reports philosophical research and does not involve human participants, animals, or identifiable personal data.

\noindent\textbf{Consent to participate.} Not applicable.

\noindent\textbf{Consent to publish.} Not applicable.

\noindent\textbf{Use of generative AI.} During revision, the authors used Anthropic Claude (Opus 4.7) and OpenAI's ChatGPT (GPT 5.6 Sol) to assist with language revision and translation. The authors reviewed and verified the resulting text and references and take full responsibility for the argument, accuracy, originality, and final wording of the manuscript.

\end{document}